\documentclass[10pt,journal]{IEEEtran}

\usepackage{amsmath,amssymb}
\usepackage{graphicx}
\usepackage{booktabs}
\usepackage{multirow}
\usepackage{cite}
\usepackage{url}
\usepackage{array}
\usepackage{tikz}
\usetikzlibrary{positioning}
\title{Contrastive Learning for Privacy Enhancements in Industrial Internet of Things}

\author{\IEEEauthorblockN{Lin Liu} \
\and 
\IEEEauthorblockN{Rita Machacy} \
\and
\IEEEauthorblockN{Simi D Kuniyilh}
}

\begin{document}
\maketitle

\begin{abstract}
The Industrial Internet of Things (IIoT) integrates intelligent sensing, communication, and analytics into industrial environments, including manufacturing, energy, and critical infrastructure. While IIoT enables predictive maintenance and cross-site optimization of modern industrial control systems, such as those in manufacturing and energy, it also introduces significant privacy and confidentiality risks due to the sensitivity of operational data. Contrastive learning, a self-supervised representation learning paradigm, has recently emerged as a promising approach for privacy-preserving analytics by reducing reliance on labeled data and raw data sharing. Although contrastive learning-based privacy-preserving techniques have been explored in the Internet of Things (IoT) domain, this paper offers a comprehensive review of these techniques specifically for privacy preservation in Industrial Internet of Things (IIoT) systems. It emphasizes the unique characteristics of industrial data, system architectures, and various application scenarios. Additionally, the paper discusses solutions and open challenges and outlines future research directions.
\end{abstract}


\section{Introduction}
The Industrial Internet of Things (IIoT) is a cornerstone of Industry~4.0, enabling interconnected machines and sensors to generate large volumes of operational data~\cite{lu2019industrial, lasi2014industry, ghobakhloo2020industry}. Although these data support intelligent industrial applications, they also pose serious security and privacy risks~\cite{mdpi2023iiotprivacy,melnyk2025hardware, li2017security}. Centralized machine learning pipelines are often incompatible with industrial privacy, intellectual property protection, and regulatory requirements~\cite{panchal2018security}.

IIoT helps enable the large-scale interconnection of sensors, controllers, machines, and cyber–physical systems to support intelligent manufacturing, predictive maintenance, and process optimization. By continuously collecting and transmitting high-dimensional operational data, IIoT systems improve efficiency and reliability but also significantly expand the attack surface for privacy leakage ~\cite{hazra2021comprehensive}.
For instance, energy meters in a factory can provide details on user or building operational patterns, schedules, machine cycles, and production volume.
In other cases, sensitive information, such as production patterns, equipment health, trade secrets, and even worker behavior, can be inferred from raw or weakly protected data streams, making privacy preservation a critical requirement alongside traditional goals of safety and availability~\cite{boyes2018industrial, yu2019survey, peng2025log}.

Conventional privacy-enhancing techniques in the Internet of Things (IoT)—such as data anonymization, encryption, access control, and differential privacy—have demonstrated effectiveness in specific IIoT scenarios. Yet, they often face limitations when deployed in heterogeneous, resource-constrained, and latency-sensitive industrial environments~\cite{panchal2018security}. Heavy cryptographic operations may introduce unacceptable delays, while strong noise injection can degrade the utility of data-driven analytics and machine learning models. At the same time, modern IIoT applications increasingly rely on data-driven intelligence at the edge and cloud, creating a pressing need for privacy-aware representation learning methods that balance data utility and confidentiality.

Contrastive learning (CL) has emerged as a self-supervised approach that focuses on representation learning rather than relying on direct access to raw data~\cite{hu2024comprehensive}. CL is a leading framework in self-supervised and representation learning and has recently proven to be a promising method for enhancing privacy in IoT systems~\cite{chathoth2025dynamic}.
By learning discriminative representations through instance-level or semantic-level comparisons rather than explicit label supervision, contrastive learning can reduce reliance on raw data sharing and enable the extraction of task-relevant yet privacy-preserving features. In IIoT settings, contrastive objectives can be designed to separate sensitive attributes from operationally useful information, support federated or distributed training, and improve robustness against inference and reconstruction attacks~\cite{hu2024comprehensive}. These properties make contrastive learning particularly well-suited to privacy-aware analytics in industrial sensing, control, and monitoring applications. Like any deep learning technique, CL is vulnerable to backdoor attacks~\cite{li2024bilevel, chathoth2025pcap, bagdasaryan2020backdoor, saha2020backdoor, chathoth2024dynamic}. Therefore, CL-based privacy-preserving techniques designed for IIoT must consider resilience against such backdoor attacks as an important design criterion for real-world deployment.

This paper investigates the role of contrastive learning in enhancing privacy for IIoT systems. We discuss the unique privacy challenges of industrial environments, analyze how contrastive learning mechanisms can be adapted to address these challenges, and highlight emerging research directions at the intersection of self-supervised learning and industrial privacy protection. By bridging recent advances in contrastive learning with the practical constraints of IIoT, this work aims to provide a foundation for developing scalable, effective, privacy-preserving intelligent industrial systems. This paper examines contrastive learning as a privacy-enabling technology for IIoT systems.

\section{background}

\subsection{Industrial IoT}
The Industrial Internet of Things (IIoT) refers to the integration of interconnected sensing, communication, and computing technologies within industrial environments to enable intelligent monitoring, control, and optimization of physical processes~\cite{xu2014iiot}. Unlike consumer-oriented IoT systems, IIoT deployments operate in mission-critical settings such as smart manufacturing, energy systems, transportation, and process industries, where reliability, real-time performance, and safety are paramount~\cite{lu2019industrial}. By leveraging distributed sensors, actuators, edge devices, and cloud platforms, IIoT enables advanced functionalities including predictive maintenance, asset tracking, digital twins, and autonomous decision-making~\cite{lasi2014industry}.

A typical IIoT architecture follows a multi-layer design comprising the perception layer (industrial sensors and actuators), the network layer (industrial Ethernet, wireless fieldbuses, and 5G), and the application layer (data analytics, control, and visualization). Recent advancements have introduced edge and fog computing layers to reduce latency, improve scalability, and support local intelligence near industrial assets. However, the heterogeneity of devices, protocols, and vendors, combined with long equipment lifecycles, makes IIoT systems complex and difficult to secure~\cite{lasi2014industry}.

From a data perspective, IIoT systems continuously generate large volumes of high-frequency, high-dimensional data that reflect operational states, production workflows, and system behaviors~\cite{lu2019industrial}. While such data are essential for industrial intelligence, they also contain sensitive information that, if exposed or improperly analyzed, can reveal proprietary processes, operational strategies, or vulnerabilities. Adversaries may exploit IIoT data through inference attacks, traffic analysis, or model-based reconstruction to extract confidential industrial knowledge~\cite{li2017security, panchal2018security}.

These characteristics distinguish IIoT from traditional IoT and motivate the need for specialized security and privacy mechanisms. In particular, privacy preservation in IIoT must balance strong protection guarantees with strict constraints on latency, computational resources, and data utility. As IIoT increasingly relies on data-driven and learning-based methods, understanding its architectural and operational context is essential for designing effective privacy-enhancing techniques, including contrastive learning–based representation learning approaches discussed in subsequent sections.

\subsection{IIoT Privacy}
Although privacy challenges and preservation techniques have been extensively looked at, the same technique can not always be applicable to IIoT~\cite{zheng2018user,abadi2016deep,chathoth2021federated,shokri2015privacy,chathoth2022differentially}. Privacy concerns and preservation in the IIoT are particularly challenging due to the scale, heterogeneity, and mission-critical nature of industrial systems~\cite{mdpi2023iiotprivacy,panchal2018security}. IIoT data streams often encode sensitive information beyond explicit measurements, including production rates, equipment utilization, control logic, and operational anomalies. Even when direct identifiers are removed, adversaries can exploit temporal correlations, statistical patterns, or side-channel information to infer confidential industrial knowledge, leading to risks such as industrial espionage, competitive disadvantage, and safety threats~\cite{mdpi2023iiotprivacy}.

One fundamental challenge arises from data inference and reconstruction attacks. Machine learning models trained on industrial data may unintentionally memorize sensitive attributes, allowing attackers to recover private information through model inversion, membership inference, or feature reconstruction attacks~\cite{li2017security}. This risk is amplified in IIoT environments where data are highly structured, periodic, and strongly correlated with physical processes. Additionally, distributed data sharing across edge devices, cloud platforms, and third-party service providers increases the risk of privacy leakage during data transmission and collaborative analytics~\cite{ghobakhloo2020industry}.

Resource and operational constraints further complicate privacy protection in IIoT. Many industrial devices have limited computational capacity and strict real-time requirements, making heavyweight cryptographic or privacy-preserving mechanisms difficult to deploy at scale. Moreover, IIoT systems often operate over long lifecycles and legacy infrastructure, limiting the feasibility of frequent updates or protocol redesigns. As a result, privacy solutions must be adaptive, lightweight, and compatible with existing industrial workflows~\cite{mdpi2023iiotprivacy}.

To address these challenges, AI-driven privacy protection mechanisms have gained increasing attention. Traditional approaches such as differential privacy (DP) have been integrated into machine learning pipelines to provide formal privacy guarantees by perturbing data or gradients during model training~\cite{chathoth2022differentially, abadi2016deep}. While effective in theory, DP-based methods often introduce a tradeoff between privacy strength and model accuracy, which can be problematic for precision-critical industrial applications~\cite{ghobakhloo2020industry}. Similarly, federated learning (FL) has been adopted to reduce raw data sharing by keeping data local to industrial sites or edge devices, but it remains vulnerable to gradient leakage and inference attacks without additional protection~\cite{mcmahan2017communication, kairouz2021federated}.

Recent research has explored representation-learning–based privacy preservation, in which AI models are designed to learn task-relevant features while suppressing or disentangling sensitive information~\cite{oord2018representation}. Adversarial learning, disentangled representation learning, and self-supervised methods have been proposed to explicitly control the privacy–utility tradeoff~\cite{hathaliya2022adversarial, xun2024cleanerclip, huang2021robust, hassan2020robust}. In this context, contrastive learning has emerged as a promising paradigm, as it enables models to learn robust and discriminative representations without direct access to labels or raw sensitive attributes~\cite{hu2024comprehensive}. By carefully constructing positive-negative pairs or by defining privacy-aware contrastive objectives, IIoT systems can reduce information leakage while maintaining high analytical performance~\cite{hu2024comprehensive}.

Overall, existing AI-based privacy protection techniques provide valuable foundations but remain insufficient to fully address the unique requirements of IIoT. The combination of strong adversaries, complex data dependencies, and stringent industrial constraints motivates the exploration of contrastive learning as a flexible and scalable approach for privacy enhancement, which is examined in detail in the following sections.

\subsection{Contrastive Learning}
Contrastive learning is a self-supervised representation learning paradigm that aims to learn informative and transferable feature embeddings by exploiting the intrinsic structure of data rather than relying on manual labels~\cite{hu2024comprehensive}. The core idea of contrastive learning is to bring semantically similar samples, referred to as positive pairs, closer together in a latent embedding space while simultaneously pushing dissimilar samples, or negative pairs, farther apart. This is typically achieved by a contrastive objective function that maximizes agreement across different augmented views of the same data instance while enforcing discrimination across instances. In practical implementations, positive pairs are often generated using domain-specific data augmentations—such as temporal cropping, noise injection, or frequency transformations in time-series data—while negative pairs are drawn from other samples within a batch or memory bank. By optimizing this objective, contrastive learning encourages models to focus on invariant and semantically meaningful features, rather than memorizing superficial or instance-specific details. This property makes contrastive learning particularly effective in scenarios where labeled data are scarce, expensive, or sensitive, as is common in large-scale sensor-driven systems. Moreover, contrastive learning has demonstrated strong generalization across downstream tasks, enabling learned representations to be reused for classification, anomaly detection, clustering, and forecasting with minimal additional supervision~\cite{hu2024comprehensive}. From a systems perspective, contrastive learning is well suited to decentralized and resource-constrained environments, as representations can be learned locally and shared in an abstract form without exposing raw data. These characteristics have made contrastive learning a foundational technique in modern self-supervised learning and a key enabler for privacy-aware, scalable, and robust machine learning across domains such as computer vision, natural language processing, and industrial time-series analytics.


\section{Privacy Challenges in Industrial IoT}
Industrial data encodes sensitive operational knowledge, including machine characteristics, fault signatures, and production strategies. Even anonymized datasets may leak proprietary information through inference attacks. These risks motivate decentralized and representation-centric learning approaches.
Industrial IoT privacy is uniquely challenging because IIoT environments generate large volumes of highly sensitive operational data that often reflect proprietary processes, machine behavior, supply-chain details, and even personnel movements. This sensitivity makes confidentiality breaches not just a data protection issue but a competitive and safety concern in industrial settings. Modern research highlights several key privacy challenges:
\begin{itemize}

\item Heterogeneity and scale of IIoT data:
IIoT systems integrate diverse sensors, actuators, controllers, and networked devices that produce heterogeneous and high-velocity data streams. The complexity and scale of these deployments make consistent privacy protection difficult. Ensuring confidentiality across multiple device types, communication protocols, and manufacturers is non-trivial, particularly in environments lacking standardized security frameworks. This heterogeneity also complicates the design of uniform privacy controls and can lead to gaps that attackers exploit
~\cite{mdpi2023iiotprivacy}.

\item Vulnerability to sophisticated attacks:
IIoT networks are susceptible to a range of advanced inference and attack vectors that threaten privacy. These include unauthorized access to industrial control systems (ICS) and Human–Machine Interfaces (HMIs), malware and ransomware targeting production infrastructure, and side-channel or traffic-analysis attacks that can infer sensitive operational patterns. Because industrial systems often cannot be taken offline, attackers have persistent opportunities to probe and exploit privacy weaknesses
~\cite{mdpi2023iiotprivacy}.

\item Difficulties ensuring confidentiality, integrity, and availability simultaneously:
Privacy in IIoT is closely tied to broader cybersecurity goals. Ensuring confidentiality while also maintaining system availability and integrity is technically complex because intensive encryption and monitoring can introduce latency or interfere with real-time control loops in industrial environments. The trade-off between robust privacy protections and real-time operational needs is a persistent technical tension
~\cite{mdpi2023iiotprivacy}.

\item Distributed and decentralized architectures:
IIoT deployments are often geographically and organizationally distributed (e.g., multiple factories or supply chain partners). Centralized data collection poses privacy risks and may violate regulatory constraints tied to cross-border data flows. However, distributed analytics raises issues of authorization, identity management, and secure sharing of aggregated information without leaking sensitive details. Maintaining trust across decentralized participants without exposing raw data remains an open problem
~\cite{arxiv2023inference}.

\item Inference and side-channel threats:
Even when raw data are not exposed, privacy can be compromised through inference attacks, in which attackers use aggregated or derived data to infer sensitive operational characteristics. This includes statistical reconstruction of production behaviors and identification of machine settings. The risk of unintended leakage through analytic models or metadata is a major concern in privacy research for IIoT~\cite{arxiv2024leakage}.

\item Regulatory and compliance complexity:
IIoT operators must often reconcile confidentiality requirements from multiple regulatory frameworks (e.g., data protection laws and industry-specific safety standards) with real-time operational requirements. Ensuring privacy without undermining compliance or system performance poses a challenging engineering and governance problem. 
~\cite{mdpi2023iiotprivacy}.

\item Lack of robust and standardized privacy mechanisms:
The literature frequently emphasizes that existing anonymization, access-control, and cryptographic techniques are often insufficient or too heavy for practical IIoT use. Passwordless devices, inconsistent policy enforcement, and under-protected communication channels create exploitable privacy gaps that persist despite research into solutions like secure enclaves, lightweight cryptography, or collaborative privacy frameworks 
~\cite{mdpi2023iiotprivacy}.
 
\end{itemize}
Privacy in Industrial Internet of Things (IIoT) systems has emerged as a critical research challenge due to the proprietary, safety-critical, and economically sensitive nature of industrial data. Unlike consumer IoT, IIoT deployments generate high-frequency sensor and control data that may encode confidential information such as machine configurations, production volumes, fault signatures, and operational strategies. Recent studies demonstrate that even anonymized or aggregated industrial data can be exploited through inference or reconstruction attacks, leading to unintended disclosure of sensitive operational knowledge \cite{shokri2015privacy,arxiv2024leakage}.

A major technical challenge identified in recent literature is the heterogeneity and non-identically distributed (non-IID) nature of IIoT data across factories, production lines, and vendors. Centralized learning approaches not only struggle with such heterogeneity but also amplify privacy risks by requiring the aggregation of raw data across organizational boundaries. Although federated learning has been proposed as a privacy-preserving alternative, recent work shows that shared gradients and model updates can still leak sensitive industrial information, motivating the need for stronger privacy-aware learning paradigms \cite{kairouz2021federated,arxiv2023inference}.

Another critical challenge is the trade-off between privacy protection and real-time operational constraints. Industrial control systems demand low latency, high availability, and deterministic behavior. Heavyweight cryptographic mechanisms or excessive noise injection, such as naive differential privacy implementations, may degrade fault detection accuracy or disrupt time-critical control loops. As highlighted in recent IIoT surveys, balancing privacy guarantees with operational reliability remains an open engineering problem \cite{mdpi2023iiotprivacy}.

Side-channel and metadata leakage further complicate the preservation of privacy in IIoT. Even when raw sensor data remains local, communication patterns, timing information, or learned representations can reveal sensitive information about production schedules or machine states. This has driven growing interest in representation-level privacy, where learning algorithms are explicitly designed to suppress sensitive attributes while retaining task-relevant information \cite{arxiv2024leakage}.

Finally, regulatory and organizational constraints exacerbate IIoT privacy challenges. Cross-border data transfer restrictions, intellectual property protection requirements, and contractual confidentiality agreements limit data sharing among industrial partners. As a result, recent research increasingly advocates for privacy-by-design learning frameworks that combine decentralized architectures, self-supervised representation learning, and formal privacy analysis to enable collaborative industrial analytics without exposing sensitive operational data \cite{xu2014iiot}.

Together, these challenges illustrate why privacy in IIoT remains a fertile research area and why approaches such as contrastive learning, federated learning, and other privacy-enhancing techniques are attracting growing attention as potential means to mitigate these risks.


\section{Contrastive Learning Techniques for Privacy Enhancements in Industrial IoT}

Contrastive learning (CL) has attracted growing attention as a privacy-preserving representation-learning paradigm for IIoT systems. Unlike supervised learning, which often requires labeled datasets containing sensitive operational information, contrastive learning enables models to learn discriminative representations by comparing similarities among data samples. This property makes CL particularly suitable for IIoT environments, where raw data exposure can lead to privacy leakage, industrial espionage, or inference of proprietary processes. This section presents an in-depth review of contrastive learning techniques for privacy preservation in IIoT, organized according to a comprehensive taxonomy and comparative analysis.


Contrastive learning seeks to maximize agreement between positive sample pairs while minimizing similarity between positive and negative pairs in a latent representation space. Contrastive learning has been widely applied in the IoT domain~\cite{chathoth2025dynamic}. Similarly, in IIoT, positive pairs may correspond to sensor readings from the same machine under similar operating conditions, while negative pairs may represent data from different machines, time periods, or operational modes. By optimizing contrastive objectives such as InfoNCE, CL models learn invariant features that capture system behavior without explicitly encoding sensitive identifiers~\cite{parulekar2023infonce}.

The relevance of contrastive learning to IIoT privacy arises from its ability to (1) reduce dependency on raw data sharing, (2) enable decentralized or federated training, and (3) support selective invariance to sensitive attributes such as device identity, production volume, or operational schedules. These properties align well with industrial requirements for confidentiality, robustness, and scalability.

\section{Taxonomy of Contrastive Learning for Privacy Preservation}

\begin{figure*}[ht]
\centering

\begin{tikzpicture}[
    node distance=1.8cm,
    every node/.style={draw, rectangle, rounded corners, align=center, font=\tiny},
    root/.style={fill=gray!15, minimum width=1.8cm, minimum height=0.55cm},
    cat/.style={fill=blue!8, minimum width=1.8cm, minimum height=0.5cm},
    sub/.style={fill=green!8, minimum width=1.8cm, minimum height=0.5cm},
    arrow/.style={->, thick}
]

\node[root] (root) {Contrastive Learning\\for IIoT Privacy};

\node[cat, below=of root, xshift=-6cm] (threats) {Privacy Threats};
\node[cat, below=of root]              (methods) {CL Techniques};
\node[cat, below=of root, xshift=6cm]  (apps) {IIoT Applications};

\node[sub, below=of threats, xshift=-2.0cm] (infer) {Inference\\Attacks};
\node[sub, below=of threats]                (leak) {Data Leakage};
\node[sub, below=of threats, xshift=2.0cm]  (poison) {Model inversion\\ /Reconstruction};

\node[sub, below=of methods, xshift=-2.0cm] (inst) {Instance-level};
\node[sub, below=of methods]                (selfsup) {Self-Supervised\\CL};
\node[sub, below=of methods, xshift=2.0cm]  (cross) {Cross-modal};

\node[sub, below=of apps, xshift=-2.0cm] (energy) {Smart Grid};
\node[sub, below=of apps]                (manufact) {Smart\\Manufacturing};
\node[sub, below=of apps, xshift=2.0cm]  (health) {Industrial\\Health};

\draw[arrow] (root.south) -- (threats.north);
\draw[arrow] (root.south) -- (methods.north);
\draw[arrow] (root.south) -- (apps.north);

\draw[arrow] (threats.south) -- (infer.north);
\draw[arrow] (threats.south) -- (leak.north);
\draw[arrow] (threats.south) -- (poison.north);

\draw[arrow] (methods.south) -- (inst.north);
\draw[arrow] (methods.south) -- (selfsup.north);
\draw[arrow] (methods.south) -- (cross.north);

\draw[arrow] (apps.south) -- (energy.north);
\draw[arrow] (apps.south) -- (manufact.north);
\draw[arrow] (apps.south) -- (health.north);

\end{tikzpicture}

\caption{Taxonomy of contrastive learning techniques for privacy enhancement in IIoT.}
\label{fig:cl_iot_taxonomy}
\end{figure*}
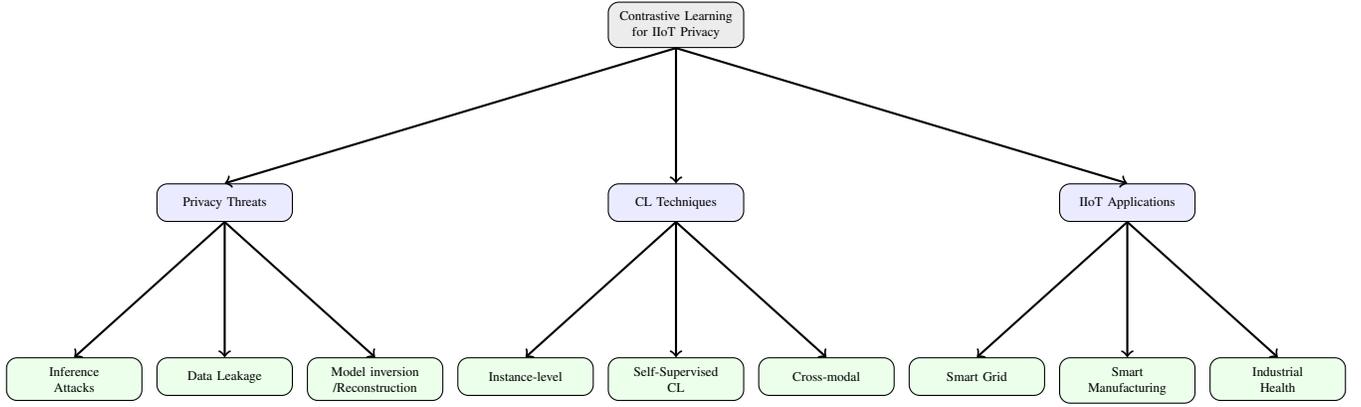

We categorize existing contrastive learning approaches for IIoT privacy across three key dimensions: privacy threats, contrastive learning techniques used, and the type of IIoT environment, as shown in Figure ~\ref{fig:cl_iot_taxonomy}. 

\subsection{Privacy Threats}
This section explains the most common privacy threats for IIoT.
\subsubsection{Inference Attacks}

Inference attacks aim to extract sensitive information about the training data, underlying system behavior, or participating devices by analyzing the learned representations or model outputs. In the context of CL for IIoT, such attacks are particularly concerning because representations are explicitly optimized to preserve semantic similarity across augmented views, which may inadvertently encode private attributes.

Membership inference attacks determine whether a specific data sample or device trace was used during training by observing model outputs or embedding distributions \cite{shokri2017membership,nasr2019comprehensive}. In industrial environments, this may reveal confidential information about production processes, machine usage, or participation of specific factories in collaborative training.

Attribute inference attacks go a step further by inferring sensitive properties of the training data, such as operating conditions, fault states, or device configurations, even when raw data are inaccessible \cite{hitaj2017deep}. In distributed or federated IIoT settings, shared contrastive representations and gradients increase exposure to such attacks, as adversaries can exploit statistical correlations embedded in the learned feature space.

\subsubsection{Data Leakage}

Data leakage refers to the unintended exposure of sensitive information through learned embeddings, intermediate features, or shared training artifacts. In CL-based IIoT systems, representations are often reused across multiple downstream tasks, including anomaly detection, predictive maintenance, and quality monitoring, amplifying the risk of privacy leakage.

Prior work has shown that deep representations may encode sufficient information to reconstruct or approximate original inputs \cite{zhu2019deep}. In IIoT scenarios, this could enable attackers to recover sensitive sensor patterns, operational cycles, or proprietary industrial signals from shared embeddings alone.

Moreover, common contrastive data augmentations—such as temporal cropping, frequency masking, or sensor fusion—do not guarantee privacy preservation. When adversaries possess auxiliary knowledge of the industrial process or sensor layout, even anonymized or encrypted embeddings may leak fine-grained operational details \cite{melis2019exploiting}. This issue is particularly critical in multi-tenant industrial platforms where representations are exchanged across organizational boundaries.

\subsubsection{Reconstruction and Model Inversion Attacks}

Reconstruction attacks, also known as model inversion attacks, aim to recover sensitive information about training samples by exploiting access to learned representations, gradients, or model parameters. In contrastive learning, embeddings are designed to be semantically rich, which can unintentionally facilitate input reconstruction.
Studies have demonstrated that attackers can reconstruct training data from shared gradients or intermediate representations in collaborative and federated learning settings \cite{zhu2019deep,geiping2020inverting}. In IIoT environments, such attacks may enable the recovery of raw sensor signals, temporal activity patterns, or proprietary industrial measurements from contrastive embeddings.

Model inversion attacks are particularly concerning in CL-based IIoT systems because embeddings are frequently transmitted to edge servers or cloud platforms for downstream analytics. Even partial access to the representation space can allow adversaries to approximate sensitive operational states or reverse-engineer machine behavior \cite{fredrikson2015model}. These threats highlight the need for embedding-level privacy protection mechanisms in contrastive representation learning.





\subsection{Contrastive Learning techniques}

Contrastive learning (CL) has emerged as a powerful paradigm for representation learning from unlabeled industrial data~\cite{hu2024comprehensive}. In IIoT environments, CL techniques enable scalable learning from heterogeneous, high-frequency sensor streams without manual annotation. However, different CL formulations exhibit distinct privacy characteristics depending on how positive and negative pairs are constructed, how representations are shared, and how multiple data modalities are integrated. This section discusses major CL techniques and their privacy implications in IIoT applications.

\subsubsection{Instance-Level Contrastive Learning}

Instance-level contrastive learning treats each individual data sample as a distinct class and learns representations by pulling together augmented views of the same instance while pushing apart representations of other instances \cite{chen2020simple}. This approach has been widely adopted due to its simplicity and strong empirical performance.

In IIoT settings, instance-level CL is commonly applied to raw sensor time series, vibration signals, or network traffic traces. From a privacy perspective, instance discrimination encourages the preservation of fine-grained, instance-specific characteristics in the learned embedding space. While this improves downstream task performance, it also increases the risk that embeddings encode sensitive operational details unique to specific machines, production lines, or industrial sites.

Furthermore, because each instance is treated as a unique identity, instance-level CL may inadvertently facilitate membership and attribute inference attacks when representations are shared across edge or cloud platforms. Without additional privacy constraints, embeddings learned via instance-level CL may remain highly linkable to individual devices or operational states.

\subsubsection{Self-Supervised Contrastive Learning}

Self-supervised contrastive learning extends the contrastive paradigm by leveraging pretext tasks and data augmentations to generate supervisory signals directly from unlabeled data \cite{he2020momentum}. In IIoT environments, self-supervised CL enables scalable learning from continuous sensor streams without costly labeling efforts.

While self-supervised CL reduces reliance on explicit annotations, it does not inherently guarantee privacy preservation. The choice of augmentations—such as temporal cropping, frequency perturbation, or signal masking—plays a critical role in determining how much sensitive information is retained in learned representations. In industrial contexts, insufficiently obfuscating augmentations may allow embeddings to preserve proprietary temporal patterns, machine signatures, or operational regimes.

Additionally, self-supervised CL models are often pretrained on large, diverse datasets and reused across multiple downstream IIoT tasks. This reuse amplifies privacy risks, as representations optimized for generality may encode latent correlations exploitable by inference or reconstruction attacks when deployed in shared or federated settings.

\subsubsection{Cross-Modal Contrastive Learning}

Cross-modal contrastive learning aligns representations across multiple data modalities by maximizing agreement between paired views from different sensors or sources \cite{radford2021learning}. In IIoT applications, this may involve correlating vibration signals with acoustic data, network telemetry with control logs, or sensor measurements with maintenance records.

From a privacy standpoint, cross-modal CL introduces unique challenges. Aligning multiple modalities can amplify information leakage, as sensitive attributes present in one modality may be transferred or inferred through another. For example, production schedules or fault conditions captured in control logs may become inferable from correlated sensor embeddings.

Moreover, cross-modal CL often requires synchronized data sharing across heterogeneous IIoT subsystems, increasing the attack surface. If adversaries gain access to one modality or its embeddings, they may exploit cross-modal alignment to infer sensitive information from other modalities, even if those data are not directly accessible. As a result, privacy risks in cross-modal CL are cumulative rather than isolated, necessitating careful design of modality alignment and representation sharing strategies.

\subsection{IIoT Applications with High Privacy Concerns}

Industrial Internet of Things (IIoT) deployments span a wide range of application domains that rely on continuous data collection from sensors, machines, and cyber-physical systems. While contrastive learning enables effective representation learning from such large-scale unlabeled data, many IIoT applications involve highly sensitive operational, economic, and safety-related information. This section highlights major IIoT application domains where privacy concerns are particularly critical.

\subsubsection{Smart Grid}

Smart grid systems integrate advanced sensing, communication, and control technologies to improve the efficiency, reliability, and sustainability of power generation and distribution. IIoT devices such as smart meters, phasor measurement units (PMUs), and substation sensors continuously collect fine-grained energy consumption and grid state data.

From a privacy perspective, smart grid data can reveal sensitive information about industrial production schedules, equipment utilization, and operational strategies. High-resolution load profiles may expose proprietary industrial processes or enable inference of business activities when analyzed over time \cite{mohassel2014survey}. When contrastive learning is applied to smart grid data for tasks such as load forecasting, anomaly detection, or fault diagnosis, learned representations may inadvertently encode these sensitive patterns.

Moreover, smart grid infrastructures often involve data sharing among utilities, operators, and third-party service providers. Transmitting contrastive embeddings to centralized platforms or cloud-based analytics services increases the risk of information leakage, particularly when embeddings are reused across multiple downstream applications \cite{efthymiou2010smart}.

\subsubsection{Smart Manufacturing}

Smart manufacturing leverages IIoT technologies to enable real-time monitoring, predictive maintenance, and adaptive control of industrial production systems. Sensors embedded in machines, production lines, and robotic systems generate continuous streams of operational data, including vibration signals, tool wear measurements, and quality indicators.

These data streams are highly sensitive, as they reflect proprietary manufacturing processes, production efficiency, and equipment performance. Contrastive learning is increasingly used to extract robust representations from unlabeled sensor data for fault detection and process optimization. However, such representations may capture fine-grained temporal and spatial characteristics unique to specific factories or production lines \cite{lee2015cyber}.

Privacy concerns arise when learned representations are shared across sites, vendors, or cloud platforms for collaborative learning or benchmarking. Even without access to raw sensor data, adversaries may infer trade secrets, production volumes, or equipment health states from contrastive embeddings, posing significant risks to industrial confidentiality and competitive advantage.

\subsubsection{Industrial Monitoring and Control Systems}

Industrial monitoring and control systems, including Supervisory Control and Data Acquisition (SCADA) systems, are responsible for maintaining safe and efficient operation of critical infrastructure such as oil refineries, chemical plants, and water treatment facilities. IIoT sensors in these environments collect real-time measurements of pressure, temperature, flow rates, and control signals.

The privacy of industrial monitoring data is closely tied to both safety and intellectual property. Detailed process measurements can reveal plant layouts, control logic, and operational thresholds. When contrastive learning is employed to model normal system behavior or detect anomalies, learned representations may encode sensitive process dynamics that are difficult to anonymize \cite{cardenas2008secure}.

Additionally, many industrial monitoring systems operate in hybrid edge–cloud architectures. Sharing contrastive representations across organizational or geographical boundaries introduces privacy risks, particularly when adversaries possess auxiliary knowledge about industrial processes. Protecting representation-level privacy is therefore essential for safe deployment of contrastive learning in industrial control environments.

\section{Contrastive Privacy Preserving Techniques}
This section presents key contrastive object designs, deployment modalities, and privacy-enhancing techniques in detail.
\subsection{Contrastive Objective Design}

\textbf{Instance-Level Contrastive Learning:}  
Instance-level CL treats each data instance as a separate class and learns representations by distinguishing one instance from all others. In IIoT, this approach is often applied to multivariate time-series sensor data, vibration signals, or energy consumption traces. While instance-level CL effectively captures fine-grained operational patterns, it may inadvertently preserve sensitive correlations, such as machine fingerprints or production rhythms, that can be exploited by inference attacks. Privacy enhancement in this category relies on carefully designed data augmentation techniques, such as temporal masking, frequency perturbation, or noise injection, to obscure sensitive patterns while preserving task relevance.

\textbf{Attribute-Aware Contrastive Learning:}  
Attribute-aware CL explicitly considers sensitive and non-sensitive attributes during representation learning. In IIoT, sensitive attributes may include factory identity, production batch information, or operational schedules. These methods construct positive pairs that share task-relevant semantics but differ in sensitive attributes, thereby encouraging invariance to private information. Often combined with adversarial objectives, attribute-aware CL penalizes representations that enable accurate prediction of sensitive attributes. This approach is particularly effective against membership and attribute inference attacks, but it requires partial knowledge or estimates of sensitive attributes.

\textbf{Temporal and Contextual Contrastive Learning:}  
IIoT data exhibit strong temporal dependencies and contextual consistency. Temporal contrastive learning exploits these properties by contrasting time-adjacent segments against temporally distant ones. Privacy benefits arise when sensitive long-term patterns are suppressed while short-term operational dynamics are preserved. Contextual CL further incorporates operational modes or environmental context to enhance robustness without revealing sensitive system-level information.

\textbf{Graph-Based Contrastive Learning:}  
Industrial systems naturally form graphs through physical connections, communication links, or functional dependencies. Graph contrastive learning leverages these structures to learn node or subgraph embeddings. Privacy is enhanced by abstracting away node-specific attributes and focusing on relational patterns. For example, device-level identifiers can be masked while preserving inter-device coordination patterns critical for fault detection and process optimization.

\subsection{Deployment Modality}

\textbf{Centralized Contrastive Learning:}  
In centralized settings, all IIoT data are aggregated for contrastive training. While this approach enables rich negative sampling and high-quality representations, it exposes sensitive data during transmission and storage. Centralized CL is generally suitable only when strong access control and encryption mechanisms are in place.

\textbf{Federated Contrastive Learning:}  
Federated CL distributes training across edge devices or industrial sites, allowing contrastive objectives to be optimized locally. Only model updates or embeddings are shared, thereby significantly reducing exposure to raw data. Federated CL aligns well with IIoT privacy requirements but introduces challenges, including non-IID data distributions, limited negative-sample diversity, and potential gradient leakage.

\textbf{Hybrid Edge--Cloud Architectures:}  
Hybrid approaches perform initial contrastive representation learning at the edge, followed by aggregation or refinement in the cloud. This balances privacy preservation with computational efficiency and scalability. However, careful task partitioning is required to prevent privacy leakage during the sharing of intermediate representations.

\subsection{Privacy Enforcement Mechanisms}
Table ~\ref{tab:privacy_enforcement} shows various privacy enforcement mechanisms in CL for IIoT.
\begin{table*}[t]
\centering
\caption{Comparison of Privacy Enforcement Mechanisms in Contrastive Learning for IIoT}
\label{tab:privacy_enforcement}
\begin{tabular}{p{3.2cm} p{3.2cm} p{2.8cm} p{3cm} p{2.2cm}}
\toprule
\textbf{Mechanism} & \textbf{Integration Strategy} & \textbf{Privacy Guarantee} & \textbf{Impact on Utility} & \textbf{IIoT Suitability} \\
\midrule
Differentially Private CL & Noise added to gradients, embeddings, or similarity scores & Formal (theoretical DP bounds) & Moderate--High degradation (tunable) & High (with careful tuning) \\
Adversarial Representation Obfuscation & Encoder trained against sensitive-attribute discriminator & Empirical (attack-dependent) & Low--Moderate degradation & High \\
Information-Theoretic Regularization & Mutual information minimization or information bottleneck constraints & Bounded information leakage & Low degradation & High \\
\bottomrule
\end{tabular}
\end{table*}

\textbf{Differentially Private Contrastive Learning:}  
Differential privacy (DP) can be integrated into CL by perturbing gradients, embeddings, or similarity scores. DP-based CL provides formal privacy guarantees against worst-case adversaries but often degrades representation quality. In IIoT, where accuracy is critical, DP parameters must be carefully tuned to balance privacy and utility.

\textbf{Adversarial Representation Obfuscation:}  
Adversarial CL frameworks introduce an auxiliary discriminator that infers sensitive attributes from learned embeddings. The encoder is trained to minimize contrastive loss while maximizing adversarial error, effectively obfuscating private information. This approach provides empirical privacy protection and is flexible but lacks formal guarantees.

\textbf{Information-Theoretic Regularization:}  
Information bottleneck and mutual information minimization techniques constrain the amount of sensitive information that can be encoded in representations. When combined with contrastive objectives, these methods explicitly limit information leakage while preserving task-relevant semantics.

\subsection{Computational Overhead and Practical Constraints}

Contrastive learning typically requires large batch sizes, extensive negative sampling, and complex encoders, which may strain IIoT edge devices. Memory banks, momentum encoders, and lightweight architectures have been proposed to reduce overhead. In federated settings, communication cost and synchronization latency further impact scalability. Therefore, privacy-enhanced CL methods must be carefully adapted to meet industrial real-time and reliability constraints.
\section{Comparative Analysis}

\begin{table*}[t]
\centering
\caption{Comparison of Privacy-Preserving Learning Approaches for IIoT}
\label{tab:comparison}
\begin{tabular}{p{3cm} p{2cm} p{2cm} p{2cm} p{2cm}}
\toprule
\textbf{Approach} & \textbf{Labels dependency} & \textbf{Raw Data Sharing} & \textbf{Privacy Level} & \textbf{IIoT Suitability} \\
\midrule
Supervised Learning & High & High & Low & Limited \\
Autoencoders & None & Medium & Medium & Moderate \\
Federated Learning & High & None & Medium--High & High \\
Differential Privacy & Varies & Varies & High & High (utility trade-off) \\
Contrastive Learning & None & None & Medium--High & High \\
Federated Contrastive Learning & None & None & High & Very High \\
\bottomrule
\end{tabular}
\end{table*}

Table~\ref{tab:comparison} presents a comparative analysis of representative learning paradigms used in IIoT systems, focusing on their reliance on labeled data, requirements for raw data sharing, achievable privacy protection, and overall suitability for industrial environments. The comparison highlights how privacy preservation in IIoT is closely intertwined with learning architecture, supervision requirements, and data locality.

Conventional supervised learning is highly dependent on labeled datasets and typically requires centralized access to raw data. While effective in controlled settings, these characteristics significantly limit its applicability in privacy-sensitive IIoT scenarios, where labeling costs are high and raw industrial data cannot be freely shared across organizational boundaries.
Autoencoder-based approaches remove the need for labels and partially reduce data exposure by learning compact latent representations. However, as shown in the table, privacy protection remains moderate since latent features may still encode sensitive operational details, particularly when representations are reused or transmitted across systems.

Federated learning improves privacy by enabling decentralized training without raw data sharing, making it well suited for distributed IIoT deployments. Nevertheless, its dependence on supervised objectives and susceptibility to inference and reconstruction attacks result in only medium to high privacy guarantees unless augmented with additional protection mechanisms.
Differential privacy provides formal, quantifiable privacy guarantees by adding noise to data, gradients, or model outputs. As shown in the table, this approach achieves strong privacy but entails an inherent trade-off between privacy and utility, which can be challenging for IIoT applications that require high accuracy and reliability.

Contrastive learning eliminates the need for labeled data and centralized raw data sharing while enabling effective representation learning from large-scale unlabeled industrial data. However, its privacy level is characterized as medium to high, as learned embeddings may still retain sensitive information in the absence of explicit privacy constraints.
Federated contrastive learning combines the advantages of decentralized training and self-supervised representation learning. By avoiding both label dependence and raw-data sharing while learning shared representations across distributed IIoT environments, it achieves strong privacy protection and high suitability for industrial applications. This integration positions federated contrastive learning as a promising foundation for future privacy-aware IIoT systems.

Overall, Table~\ref{tab:comparison} illustrates that no single approach universally satisfies all IIoT requirements. Instead, the most effective privacy-preserving solutions arise from combining decentralized learning, representation-based objectives, and formal privacy mechanisms, with federated contrastive learning emerging as a particularly compelling direction.

\begin{table*}[t]
\caption{Comparison of Contrastive Learning Techniques for Privacy Enhancement in IIoT}
\label{tab:contrastive_comparison}
\centering
\begin{tabular}{lcccc}
\hline
\textbf{Method Type} & \textbf{Objective Design} & \textbf{Privacy Strength} & \textbf{Deployment} & \textbf{Overhead} \\
\hline
Instance-level CL & Instance discrimination & Low & Centralized & Moderate \\
Attribute-aware CL & Sensitive disentanglement & Medium--High & Federated/Hybrid & High \\
Temporal CL & Time-invariant learning & Medium & Edge/Hybrid & Moderate \\
Graph-based CL & Structural contrast & Medium--High & Edge/Hybrid & High \\
DP-CL & Noise-perturbed contrast & Strong & Centralized/Federated & Very High \\
Adversarial CL & Obfuscation via discriminator & Strong & Federated & Very High \\
\hline
\end{tabular}
\end{table*}

Table~\ref{tab:contrastive_comparison} compares representative contrastive learning techniques with respect to their objective design, privacy protection capability, deployment characteristics, and computational or communication overhead in IIoT environments. The comparison highlights that privacy enhancement in CL is achieved not by a single mechanism but by a combination of representation design, training architecture, and privacy-enforcement strategies.
Instance-level contrastive learning prioritizes instance discrimination and representation separability, resulting in strong feature expressiveness but limited privacy protection. As shown in the table, its centralized deployment and moderate overhead make it suitable for controlled environments, but the lack of explicit privacy constraints exposes learned embeddings to inference and reconstruction attacks in IIoT settings.

Attribute-aware contrastive learning explicitly incorporates sensitive attribute disentanglement into the contrastive objective. By separating privacy-sensitive factors from task-relevant representations, this approach achieves medium to high privacy strength. However, the added complexity of attribute modeling and coordination across participants leads to higher overhead and favors federated or hybrid deployment scenarios.

Temporal contrastive learning focuses on learning time-invariant representations from sequential industrial data. This objective provides moderate privacy protection by reducing sensitivity to transient or fine-grained temporal patterns, while maintaining relatively low overhead. Its compatibility with edge and hybrid deployments makes it attractive for latency-sensitive IIoT applications.

Graph-based contrastive learning captures relational and structural information among devices, sensors, or system components. While this enables robust representation learning in complex industrial systems, structural embeddings may still leak sensitive topology or interaction patterns. Consequently, graph-based CL offers medium to high privacy protection at the cost of increased computational and communication overhead.

Differentially private contrastive learning (DP-CL) integrates formal privacy guarantees by injecting noise into the contrastive objective, gradients, or representations. As reflected in the table, DP-CL provides strong privacy protection but incurs very high overhead and potential utility degradation, requiring careful tuning for practical IIoT deployments.

Adversarial contrastive learning enhances privacy by introducing discriminators that explicitly prevent sensitive information leakage from learned embeddings. This approach achieves strong privacy protection but significantly increases training complexity and overhead, making it most suitable for federated IIoT environments where privacy requirements outweigh computational costs.

Overall, Table~\ref{tab:contrastive_comparison} demonstrates a fundamental trade-off between privacy strength and system overhead. Techniques that provide strong privacy guarantees typically require more complex objectives and higher resource consumption, whereas lightweight contrastive methods offer better scalability but weaker privacy protection. These trade-offs should be carefully considered when selecting CL techniques for privacy-critical IIoT applications.

\section{Discussion and Open Challenges}
Key challenges include designing industrially meaningful augmentations, quantifying privacy leakage from embeddings, and developing lightweight contrastive models for edge devices.
While contrastive learning offers a promising approach to privacy-preserving IIoT analytics, several challenges remain. Designing privacy-aware positive and negative pairs without explicit sensitive labels is non-trivial. Federated contrastive learning suffers from limited negative diversity and communication constraints. Moreover, most existing approaches lack formal privacy guarantees or comprehensive evaluation against industrial threat models. Addressing these challenges requires interdisciplinary efforts combining self-supervised learning, industrial control theory, and privacy engineering.

\section{Summary}
Contrastive learning provides a promising foundation for privacy-preserving techniques in IIoT systems. Its self-supervised and representation-centric nature aligns well with industrial privacy and deployment constraints.
In summary, contrastive learning offers a flexible and scalable framework for enhancing privacy in IIoT systems. Through carefully designed objectives, deployment strategies, and privacy-enforcement mechanisms, CL can significantly reduce the leakage of sensitive information while maintaining high analytical performance. However, achieving practical adoption in real-world industrial environments requires further research on efficiency, robustness, and formal privacy assurances.

\bibliographystyle{IEEEtran}
\bibliography{bib}

\end{document}